\documentclass[sigconf]{acmart}
\usepackage{balance}
\def\BibTeX{{\rm B\kern-.05em{\sc i\kern-.025em b}\kern-.08emT\kern-.1667em\lower.7ex\hbox{E}\kern-.125emX}}

\copyrightyear{2019}
\acmYear{2019}
\setcopyright{acmcopyright}
\acmConference[CIKM '19]{The 28th ACM International Conference on Information and Knowledge Management}{November 3--7, 2019}{Beijing, China}
\acmBooktitle{The 28th ACM International Conference on Information and Knowledge Management (CIKM '19), November 3--7, 2019, Beijing, China}
\acmPrice{15.00}
\acmDOI{10.1145/3357384.3358140}
\acmISBN{978-1-4503-6976-3/19/11}

\usepackage{multirow}
\begin{document}

\fancyhead{}
  % do not delete this code.

% The "title" command has an optional parameter, allowing the author to define a "short title" to be used in page headers.
\title{Interactive Matching Network for Multi-Turn Response Selection in Retrieval-Based Chatbots}

\author{Jia-Chen Gu$^1$, Zhen-Hua Ling$^1$, Quan Liu$^{1,2}$}
\affiliation{\institution{$^1$University of Science and Technology of China, Hefei, China}}
\affiliation{\institution{$^2$State Key Laboratory of Cognitive Intelligence, iFLYTEK Research, Hefei, China}}
\email{gujc@mail.ustc.edu.cn, zhling@ustc.edu.cn, quanliu@ustc.edu.cn}

%
% By default, the full list of authors will be used in the page headers. Often, this list is too long, and will overlap
% other information printed in the page headers. This command allows the author to define a more concise list
% of authors' names for this purpose.

%
% The abstract is a short summary of the work to be presented in the article.
\begin{abstract}
  In this paper, we propose an interactive matching network (IMN) for the multi-turn response selection task. First, IMN constructs word representations from three aspects to address the challenge of out-of-vocabulary (OOV) words. Second, an attentive hierarchical recurrent encoder (AHRE), which is capable of encoding sentences hierarchically and generating more descriptive representations by aggregating with an attention mechanism, is designed. Finally, the bidirectional interactions between whole multi-turn contexts and response candidates are calculated to derive the matching information between them. Experiments on four public datasets show that IMN outperforms the baseline models on all metrics, achieving a new state-of-the-art performance and demonstrating compatibility across domains for multi-turn response selection.
\end{abstract}


%
% Keywords. The author(s) should pick words that accurately describe the work being
% presented. Separate the keywords with commas.

%
% A "teaser" image appears between the author and affiliation information and the body
% of the document, and typically spans the page.
%  \Description{Enjoying the baseball game from the third-base seats. Ichiro Suzuki preparing to bat.}

%
% This command processes the author and affiliation and title information and builds
% the first part of the formatted document.
\maketitle

\section{Introduction}
  Building a chatbot that can converse naturally with humans on open domain topics is a challenging yet intriguing problem in artificial intelligence \cite{DBLP:journals/sigkdd/ChenLYT17}. Response selection, which aims to select the best-matched response from a set of candidates given the context of a conversation, is an important retrieval-based approach for chatbots \cite{DBLP:conf/sigdial/LowePSP15,DBLP:conf/acl/WuWXZL17,DBLP:conf/acl/WuLCZDYZL18}.

  The techniques of word embeddings and sentence embeddings are important to response selection as well as many other natural language processing (NLP) tasks. The context and the response must be projected to a vector space appropriately to capture their relationships, which are essential for the subsequent procedures. Typically, word embeddings established on the task-specific training set and a single-layer recurrent neural network are employed for the response selection task. Another key technique to the response selection task lies in context-response matching. \citet{DBLP:conf/acl/ChenZLWJI17} showed that interactions between pairs of sentences can provide useful information to help matching.

  \citet{DBLP:conf/acl/WuWXZL17} proposed the sequential matching network (SMN) to match the response with each utterance and then to accumulate matching information by an RNN. \citet{DBLP:conf/coling/ZhangLZZL18} refined utterance and employed self-matching attention to route the vital information in each utterance based on the SMN. \citet{DBLP:conf/acl/WuLCZDYZL18} proposed the deep attention matching network (DAM) to construct representations at different granularities with stacked self-attention.

  In this paper, we propose a novel neural network architecture, called the interactive matching network (IMN), for multi-turn response selection in retrieval-based chatbots. Our proposed IMN is similar to SMN but has three main differences: (1) constructing word representations from three aspects to enhance the representations at the word-level, (2) enhancing sentence representations through  an attentive hierarchical recurrent encoder to enhance the representations at the sentence-level and (3) capturing interactions between contexts and responses by collecting matching information bidirectionally to enrich the representations.

  We test our model on Ubuntu Dialogue Corpus V1 \cite{DBLP:conf/sigdial/LowePSP15}, Ubuntu Dialogue Corpus V2 \cite{DBLP:journals/dad/LowePSCLP17}, Douban Conversation Corpus \cite{DBLP:conf/acl/WuWXZL17} and E-commerce Dialogue Corpus \cite{DBLP:conf/coling/ZhangLZZL18}. The results show that our model can outperform the baseline models on all metrics, achieving new state-of-the-art performance and showing compatibility across domains for multi-turn response selection.

  In summary, our contributions in this paper are threefold. (1) This paper proposes a new model, named IMN, for multi-turn response selection in retrieval-based chatbots. (2) The empirical results show that our proposed model outperforms the baseline models in terms of all metrics on four datasets, achieving new state-of-the-art performance for multi-turn response selection. (3) This paper presents detailed experiments and discussions on contributions of each part to context-response pair matching.

  \begin{figure*}
    \centering
    \includegraphics[width=15cm]{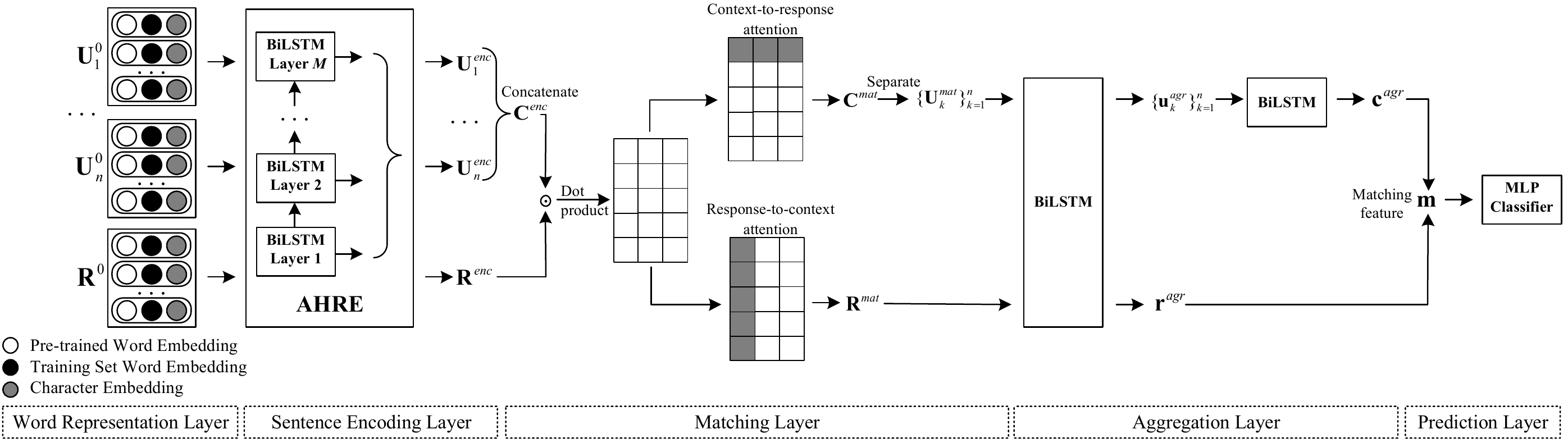}
    \caption{An overview of our proposed IMN model.}
    \label{fig1}
  \end{figure*}

\section{Interactive Matching Network}

  We present here our proposed IMN model, which is composed of five layers.% word representation layer, sentence encoding layer, matching layer, aggregation layer and prediction layer.
  Figure~\ref{fig1} shows an overview of the architecture. %Details about each layer are provided in the following sections.

  \subsection{Problem Formalization}
    Given a dialogue dataset $\mathcal{D}$, an example of the dataset can be represented as $(c,r,y)$. Specifically, $c = \{u_1,u_2,...,u_n\}$ represents a conversation context with $\{u_k\}_{k=1}^n$ as the utterances. $r$ is a response candidate, and $y \in \{0,1\}$ denotes a label. $y=1$ indicates that $r$ is a proper response for $c$; otherwise, $y=0$. Our goal is to learn a matching model $g(c,r)$, which provides the matching degree between $c$ and $r$ by minimizing the sigmoid cross entropy from $\mathcal{D}$.
%    Let $\Theta$ denote the parameters of IMN; then, the objective function $\mathcal{L}(\mathcal{D}, \Theta)$ of learning can be formulated as
%    \begin{equation}
%    \begin{aligned}
%    \mathcal{L}(\mathcal{D}, \Theta) = & - \sum_{(c,r,y)\in \mathcal{D}} [ y log(g(c,r))  \\
%                                       & + (1-y)log(1-g(c,r)) ]
%    \label{equ4}
%    \end{aligned}
%    \end{equation}

  \subsection{Word Representation Layer}

    One challenge of large dialogue corpora is the large number of OOV words. To address this issue, we propose to construct word representations with a combination of general pretrained word embedding, those estimated on the task-specific training set and character-level embeddings. %Readers can refer to Appendix A.2 for more details.

    Formally, the embeddings of the \emph{k}-th utterance in a conversation and a response candidate at this layer are denoted as $\textbf{U}_k^0 = \{\textbf{u}_{k,i}^0\}_{i=1}^{l_{u_k}}$ and $\textbf{R}^0 = \{\textbf{r}_j^0\}_{j=1}^{l_r}$. $\textbf{u}_{k,i}^0$ and $\textbf{r}_j^0 \in \mathbb{R}^d$ are embeddings of a \emph{d}-dimensional vector. ${l_{u_k}}$ and ${l_r}$ are the numbers of words in $\textbf{U}_k^0$ and $\textbf{R}^0$ respectively.

  \subsection{Sentence Encoding Layer}

    Typically, the outputs of the top layer in a multi-layer RNNs are regarded as the final sentence representations, and the other layers are neglected. However, the lower layers can also provide useful sentence descriptions, such as part-of-speech tagging and syntax-related information. Motivated by the method of ELMo \cite{DBLP:conf/naacl/PetersNIGCLZ18}, we propose a new sentence encoder, called the attentive hierarchical recurrent encoder (AHRE) to make full use of the representations at all hidden layers.

    BiLSTMs \cite{DBLP:journals/neco/HochreiterS97} are employed as our basic building blocks. In an \emph{M}-layer RNN, each $m^{th}$ layer takes the output of the ${m-1}^{th}$ layer as its input.

    Finally, we obtain a set of \emph{M} representations \{$\textbf{U}_k^{1}, ..., \textbf{U}_k^{M}$\} and \{$\textbf{R}^{1}, ..., \textbf{R}^{M}$\} for the \emph{k}-th utterance in a conversation and a response candidate through the \emph{M}-layer RNNs, where $\textbf{U}_k^m = \{\textbf{u}_{k,i}^m\}_{i=1}^{l_{u_k}}$ and $\textbf{R}^m = \{\textbf{r}_j^m\}_{j=1}^{l_r}$, $l \in \{1, ..., M\}$.
    %Typically, $\textbf{U}_k^{L}$ or $\textbf{R}^{L}$, i.e., the outputs of the top layer, are used as the final encoded vectors.
    Here, we propose to combine the set of representations to obtain the enhanced representations $\textbf{u}_{k,i}^{enc}$ and $\textbf{r}_j^{enc}$ by learning the attention weights of all the layers. Mathematically, we have
%    \begin{align}
%    \textbf{u}_{k,i}^{enc} = \sum_{l=1}^L w_l\textbf{u}_{k,i}^{l}, i \in \{1, ..., l_{u_k}\},\\
%    \textbf{r}_j^{enc} = \sum_{l=1}^L w_l\textbf{r}_j^{l}, j \in \{1, ..., l_r\},
%    \end{align}
    \begin{align}
    \textbf{u}_{k,i}^{enc} = \sum_{m=1}^M w_m\textbf{u}_{k,i}^m,~
    ~\textbf{r}_j^{enc} = \sum_{m=1}^M w_m\textbf{r}_j^m,
    \end{align}
    where $\textbf{U}_k^{enc} = \{\textbf{u}_{k,i}^{enc}\}_{i=1}^{l_{u_k}}$, $\textbf{R}^{enc} = \{\textbf{r}_j^{enc}\}_{j=1}^{l_r}$ and $w_l$ are the softmax-normalized weights shared between utterances and responses, which need to be estimated during the training process. As a result, representations given by AHRE are expected to fuse multi-level characteristics of sentences.

  \subsection{Matching Layer}

    \begin{table*}[!hbt]
     \small
     \caption{Evaluation results of IMN and previous methods on Ubuntu Dialogue Corpus V1 and  V2.}
     \centering
     \begin{tabular}{l|c|c|c|c|c|c|c|c}
      \toprule
                             & \multicolumn{4}{c|}{Ubuntu Corpus V1} & \multicolumn{4}{c}{Ubuntu Corpus V2} \\
      \hline
                             & $\textbf{R}_2@1$ & $\textbf{R}_{10}@1 $ & $\textbf{R}_{10}@2 $ & $\textbf{R}_{10}@5 $ & $\textbf{R}_2@1$ & $\textbf{R}_{10}@1 $ & $\textbf{R}_{10}@2 $ & $\textbf{R}_{10}@5 $\\
      \hline
%       TF-IDF \cite{DBLP:conf/sigdial/LowePSP15,DBLP:journals/dad/LowePSCLP17}    & 0.659 & 0.410 & 0.545 & 0.708 & 0.749 & 0.488 & 0.587 & 0.763 \\
%       RNN \cite{DBLP:conf/sigdial/LowePSP15,DBLP:journals/dad/LowePSCLP17}       & 0.768 & 0.403 & 0.547 & 0.819 & 0.777 & 0.379 & 0.561 & 0.836 \\
%       LSTM \cite{DBLP:conf/sigdial/LowePSP15,DBLP:journals/dad/LowePSCLP17}      & 0.878 & 0.604 & 0.745 & 0.926 & 0.869 & 0.552 & 0.721 & 0.924 \\

%       DL2R \cite{DBLP:conf/sigir/YanSW16}                & 0.899 & 0.626 & 0.783 & 0.944 & -     & -     & -     & -     \\
%       Match-LSTM \cite{DBLP:conf/naacl/WangJ16}          & 0.904 & 0.653 & 0.799 & 0.944 & -     & -     & -     & -     \\
%       MV-LSTM \cite{DBLP:conf/ijcai/WanLXGPC16}          & 0.906 & 0.653 & 0.804 & 0.946 & -     & -     & -     & -     \\
%       Multi-View \cite{DBLP:conf/emnlp/ZhouDWZYTLY16}    & 0.908 & 0.662 & 0.801 & 0.951 & -     & -     & -     & -     \\
%       RNN-CNN \cite{DBLP:journals/corr/BaudisS16}        & -     & -     & -     & -     & 0.911 & 0.672 & 0.809 & 0.956 \\
       CompAgg \cite{DBLP:journals/corr/WangJ16b}         & 0.884 & 0.631 & 0.753 & 0.927 & 0.895 & 0.641 & 0.776 & 0.937 \\
       BiMPM \cite{DBLP:conf/ijcai/WangHF17}              & 0.897 & 0.665 & 0.786 & 0.938 & 0.877 & 0.611 & 0.747 & 0.921 \\
       HRDE-LTC \cite{DBLP:conf/naacl/YoonSJ18}           & 0.916 & 0.684 & 0.822 & 0.960 & 0.915 & 0.652 & 0.815 & 0.966 \\
       SMN \cite{DBLP:conf/acl/WuWXZL17}                  & 0.926 & 0.726 & 0.847 & 0.961 & -     & -     & -     & -     \\
       DUA \cite{DBLP:conf/coling/ZhangLZZL18}            & -     & 0.752 & 0.868 & 0.962 & -     & -     & -     & -     \\
       DAM \cite{DBLP:conf/acl/WuLCZDYZL18}               & 0.938 & 0.767 & 0.874 & 0.969 & -     & -     & -     & -     \\
      \hline
       IMN                  & \textbf{0.946} & \textbf{0.794} & \textbf{0.889} & \textbf{0.974} & \textbf{0.945} & \textbf{0.771} & \textbf{0.886} & \textbf{0.979} \\
       IMN(Ensemble)        & \textbf{0.951} & \textbf{0.807} & \textbf{0.900} & \textbf{0.978} & \textbf{0.950} & \textbf{0.791} & \textbf{0.899} & \textbf{0.982}  \\
      \bottomrule
      \end{tabular}
      \label{tab2}
    \end{table*}

    \begin{table*}[!hbt]
     \small
     \caption{Evaluation results of IMN and previous methods on the Douban Conversation Corpus and E-commerce Corpus.}
     %All the results except ours are copied from \citet{DBLP:conf/acl/WuWXZL17,DBLP:conf/coling/ZhangLZZL18,DBLP:conf/acl/WuLCZDYZL18}.
     \centering
     \begin{tabular}{l|c|c|c|c|c|c|c|c|c}
      \toprule
                             & \multicolumn{6}{c|}{Douban Conversation Corpus} & \multicolumn{3}{c}{E-commerce Corpus} \\
      \hline
                             & \textbf{MAP} & \textbf{MRR} & $\textbf{P}@1$ & $\textbf{R}_{10}@1 $ & $\textbf{R}_{10}@2 $ & $\textbf{R}_{10}@5 $ & $\textbf{R}_{10}@1 $ & $\textbf{R}_{10}@2 $ & $\textbf{R}_{10}@5 $ \\
      \hline
%       TF-IDF                & 0.331 & 0.359 & 0.180 & 0.096 & 0.172 & 0.405 & 0.159 & 0.256 & 0.477  \\
%       RNN                   & 0.390 & 0.422 & 0.208 & 0.118 & 0.223 & 0.589 & 0.325 & 0.463 & 0.775  \\
%       LSTM                  & 0.485 & 0.527 & 0.320 & 0.187 & 0.343 & 0.720 & 0.365 & 0.536 & 0.828  \\
%       Multi-View            & 0.505 & 0.543 & 0.342 & 0.202 & 0.350 & 0.729 & 0.421 & 0.601 & 0.861  \\
       %DL2R                  & 0.488 & 0.527 & 0.330 & 0.193 & 0.342 & 0.705 & 0.399 & 0.571 & 0.842  \\
%       MV-LSTM               & 0.498 & 0.538 & 0.348 & 0.202 & 0.351 & 0.710 & 0.412 & 0.591 & 0.857  \\
%       Match-LSTM            & 0.500 & 0.537 & 0.345 & 0.202 & 0.348 & 0.720 & 0.410 & 0.590 & 0.858  \\
       SMN \cite{DBLP:conf/acl/WuWXZL17}                  & 0.529 & 0.569 & 0.397 & 0.233 & 0.396 & 0.724 & 0.453 & 0.654 & 0.886  \\
       DUA \cite{DBLP:conf/coling/ZhangLZZL18}            & 0.551 & 0.599 & 0.421 & 0.243 & 0.421 & 0.780 & 0.501 & 0.700 & 0.921  \\
       DAM \cite{DBLP:conf/acl/WuLCZDYZL18}               & 0.550 & 0.601 & 0.427 & 0.254 & 0.410 & 0.757 & -     & -     & -      \\
      \hline
       IMN                  & \textbf{0.570} & \textbf{0.615} & \textbf{0.433} & \textbf{0.262} & \textbf{0.452} & \textbf{0.789} & \textbf{0.621} & \textbf{0.797} & \textbf{0.964}\\
       IMN(Ensemble)        & \textbf{0.576} & \textbf{0.618} & \textbf{0.441} & \textbf{0.268} & \textbf{0.458} & \textbf{0.796} & \textbf{0.672} & \textbf{0.845} & \textbf{0.970} \\
      \bottomrule
      \end{tabular}
      \label{tab3}
    \end{table*}

    %Interactions between the context and the response provide important information for deciding the matching degree between them.
    Unlike previous work, which matches responses with each utterance in a context separately in an utterance-response manner \cite{DBLP:conf/acl/WuWXZL17,DBLP:conf/acl/WuLCZDYZL18,DBLP:conf/coling/ZhangLZZL18}, IMN matches the response with the whole context in a global context-response way, i.e., considering the whole context as a single sequence.
    %The goal of utterance-to-response matching is to collect the relevant parts in each utterance while neglecting the possible premise that the whole utterance is irrelevant to the response. Collecting any part of an irrelevant utterance introduces noise for the matching process. Instead,
    The global context-response matching can help select the most relevant parts of the whole context and neglect the irrelevant parts.

    First, the context $\textbf{C}^{enc} = \{\textbf{c}_i^{enc}\}_{i=1}^{l_c} $ with $l_c = \sum_{k=1}^{n} l_{u_k}$ is formed by concatenating the set of utterance representations $\{\textbf{U}_{k}^{enc}\}_{k=1}^{n}$.

    Then, an attention-based alignment is employed to collect information between two sequences by computing the attention weight between each tuple as $e_{ij} = (\textbf{c}_i^{enc})^T \cdot \textbf{r}_j^{enc}$.%\{$\textbf{c}_i^{enc}, \textbf{r}_j^{enc}$\}

    %Furthermore, local inference is determined by the attention weights $e_{ij}$ computed above to obtain the local relevance between a context and a response bidirectionally.
    For a word in the response, its response-to-context relevant representation is composed as
    \begin{equation}
    \bar{\textbf{r}}_j^{enc} = \sum_{i=1}^{l_c} \frac{exp(e_{ij})} {\sum_{k=1}^{l_c} exp(e_{kj})} \textbf{c}_i^{enc}, j \in \{1, ..., l_r\},
    \label{equ3}
    %\bar{\textbf{c}}_i^{enc} = \sum_{j=1}^{l_r} \frac{exp(e_{ij})} {\sum_{k=1}^{l_r} exp(e_{ik})} \textbf{r}_j^{enc}, i \in \{1, ..., l_c\},
    %\label{equ2}
    \end{equation}
    where $\bar{\textbf{R}}^{enc} = \{\bar{\textbf{r}}_j^{enc}\}_{j=1}^{l_r}$, $\bar{\textbf{r}}_j^{enc}$ is a weighted summation of $\{\textbf{c}_i^{enc}\}_{i=1}^{l_c}$.
    %Intuitively, the contents in $\{\textbf{r}_j^{enc}\}_{j=1}^{l_r}$ that are relevant to $\textbf{c}_i^{enc}$ are selected to form $\bar{\textbf{c}}_i^{enc}$.
    The same calculation is performed for each word in a context to form context-to-response representation $\bar{\textbf{C}}^{enc} = \{\bar{\textbf{c}}_i^{enc}\}_{i=1}^{l_c}$.
%    \begin{equation}
%    \bar{\textbf{r}}_j^{enc} = \sum_{i=1}^{l_c} \frac{exp(e_{ij})} {\sum_{k=1}^{l_c} exp(e_{kj})} \textbf{c}_i^{enc}, j \in \{1, ..., l_r\},
%    \label{equ3}
%    \end{equation}

    %To further enhance the collected information, we compute the differences and the element-wise products between \{$\textbf{C}^{enc}, \bar{\textbf{C}}^{enc}$\} and between \{$\textbf{R}^{enc}, \bar{\textbf{R}}^{enc}$\}. The differences and element-wise products are then concatenated with the original vectors to obtain the enhanced representations, as follows,
    To further enhance the collected information, the matching matrices are formed as
    \begin{align}
    \textbf{C}^{mat} &= [\textbf{C}^{enc}; \bar{\textbf{C}}^{enc}; \textbf{C}^{enc} - \bar{\textbf{C}}^{enc} ;\textbf{C}^{enc} \odot \bar{\textbf{C}}^{enc}],\\
    \textbf{R}^{mat} &= [\textbf{R}^{enc}; \bar{\textbf{R}}^{enc}; \textbf{R}^{enc} - \bar{\textbf{R}}^{enc} ;\textbf{R}^{enc} \odot \bar{\textbf{R}}^{enc}].
    \end{align}

    Finally, the concatenated context $\textbf{C}^{mat}$ need to be converted to separate utterances $\{\textbf{U}_k^{mat}\}_{k=1}^n$.

  \subsection{Aggregation Layer}
    The aggregation layer converts the matching matrices of separated utterances and responses into a final matching vector.
    %Here, a BiLSTM and a combination of max pooling and last hidden state pooling are employed to obtain the utterance and response embeddings.

    First, the set of utterance embeddings $\textbf{U}^{agr} = \{\textbf{u}_{k}^{agr}\}_{k=1}^n$ and the response embeddings $\textbf{r}^{agr}$ are obtained by composing the enhanced local matching information $\textbf{U}_k^{mat}$ and $\textbf{R}^{mat}$ with a BiLSTM, and a combination of max pooling and last hidden state pooling.

    Furthermore, the set of utterance inference vectors $\textbf{U}^{agr} = \{\textbf{u}_{k}^{agr}\}_{k=1}^n$ is fed into another BiLSTM in chronological order of the utterances in the context, followed by another pooling operation to obtain the aggregated context embeddings $\textbf{c}^{agr}$.

    The final matching feature vector is the concatenation of the context embeddings and the response embeddings as $\textbf{m} = [\textbf{c}^{agr};\textbf{r}^{agr}]$.

  \subsection{Prediction Layer}

    We then input the matching feature vector $\textbf{m}$ into a multi-layer perceptron (MLP) classifier. The MLP returns a score to denote the matching degree of a context-response pair.

\section{Experiments}

  \subsection{Datasets}

    We tested IMN on Ubuntu Dialogue Corpus V1 \cite{DBLP:conf/sigdial/LowePSP15}, Ubuntu Dialogue Corpus V2 \cite{DBLP:journals/dad/LowePSCLP17}, Douban Conversation Corpus \cite{DBLP:conf/acl/WuWXZL17} and E-commerce Dialogue Corpus \cite{DBLP:conf/coling/ZhangLZZL18}.
    %We adopted the version of Ubuntu V1 shared in \citet{DBLP:journals/corr/XuLWSW16}, in which numbers, paths and URLs were replaced by placeholders.
    %Readers can refer to Appendix A.1 for more details about the datasets.

  \subsection{Evaluation Metrics}
    We used the same evaluation metrics as those used in previous work \cite{DBLP:conf/sigdial/LowePSP15,DBLP:conf/acl/WuWXZL17,DBLP:conf/coling/ZhangLZZL18}. We calculated the recall of the true positive replies among the $k$ selected responses from $n$ available candidates, denoted as $\textbf{R}_n@k$. In addition, mean average precision (\textbf{MAP}), mean reciprocal rank (\textbf{MRR}) and precision-at-one ($\textbf{P}@1$), are especially considered for the Douban corpus, following the settings of previous work.

  \subsection{Experimental Results}

    Table~\ref{tab2} and Table~\ref{tab3} present the evaluation results of IMN and previous methods. All the results except ours are from the existing literature. IMN  outperforms other models on all metrics and datasets, which demonstrates its ability to select the best-matched response and its compatibility across domains (system troubleshooting, social network and e-commerce). The Douban Corpus includes multiple correct candidates for a context in its test set. Hence, $\textbf{MAP}$ and $\textbf{MRR}$ are recommended for reference.

    %Our proposed model outperforms the baseline model SMN by a large margin of 6.7\% in terms of $\textbf{R}_{10}@1$ on Ubuntu Dialogue Corpus V1; 4.1\% in terms of \textbf{MAP} and 4.6\% in terms of \textbf{MRR} on the Douban Conversation Corpus; and 16.8\% in terms of $\textbf{R}_{10}@1$ on the E-commerce Dialogue Corpus. Moreover,
    Our proposed model outperforms the present state-of-the-art methods on the respective datasets by a margin of 2.6\% in terms of $\textbf{R}_{10}@1$ on Ubuntu V1; 11.9\% in terms of $\textbf{R}_{10}@1$ on Ubuntu V2; 2.0\% in terms of \textbf{MAP} and 1.4\% in terms of \textbf{MRR} on Douban Corpus; and 12.0\% in terms of $\textbf{R}_{10}@1$ on E-commerce Corpus, achieving a new state-of-the-art performance on all datasets. Furthermore, we provide ensemble models built by averaging the outputs of four single models with identical architectures and different random initializations.
    Our code has been published at \emph{https://github.com/JasonForJoy/IMN} to help replicate our results. %~\footnote{https://github.com/JasonForJoy/IMN}.

\section{Ablations and Analysis}

    To demonstrate the importance of each component in our proposed model, various parts of the architecture were ablated, as shown in Table~\ref{tab4}.

    \begin{table}
     \small
     \caption{Ablation tests on Ubuntu V2 test set.}
     \centering
     \begin{tabular}{lcccc}
      \toprule
      %                       & \multicolumn{4}{c}{Ubuntu Corpus V2} \\
                             & $\textbf{R}_2@1$ & $\textbf{R}_{10}@1 $ & $\textbf{R}_{10}@2 $ & $\textbf{R}_{10}@5 $ \\
      \midrule
       IMN                   & 0.945 & 0.771 & 0.886 & 0.979  \\
       - AHRE                & 0.940 & 0.758 & 0.874 & 0.974  \\
       - Char emb            & 0.941 & 0.762 & 0.878 & 0.976  \\
       - Match               & 0.904 & 0.613 & 0.792 & 0.958  \\
      \bottomrule
      \end{tabular}
      \label{tab4}
    \end{table}

    \paragraph{AHRE} %The number of layers in the AHRE was tuned on the validation set of Ubuntu V2, as shown in Figure~\ref{fig2}, and
    The number of layers in the AHRE was set to 3. The AHRE can be considered as a generalized recurrent encoder that degenerates into a single-layer RNN when the number of layers in the AHRE is set to 1. The softmax-normalized weights of layers in the AHRE are listed in Table~\ref{tab5}, which indicates that each layer of the multi-layer RNNs contributes to the embeddings. %Readers can refer to Appendix A.3 for details.

    \begin{table}
     %\small
     \caption{Layer-wise weights of a three-layer AHRE.}
     \centering
     \begin{tabular}{lccc}
      \toprule
                  & Layer 1 & Layer 2 & Layer 3 \\
      \midrule
       Weights    & 0.4938  & 0.2181  & 0.2881  \\
      \bottomrule
      \end{tabular}
      \label{tab5}
    \end{table}

    \paragraph{Char emb} The character embeddings in the word representation layer were ablated, which resulted in a performance decrease. Additionally, we found that the lowest layer of the RNN in the AHRE constituted the highest weight, as shown in Table~\ref{tab5}. These two results may be explained by the importance of morphology information to the response selection.
    %i.e., morphology information may help to match similar words literally.
    %A response with more literally similar words may be more appropriate.

\begin{figure}
    \centering
    \includegraphics[width=5cm]{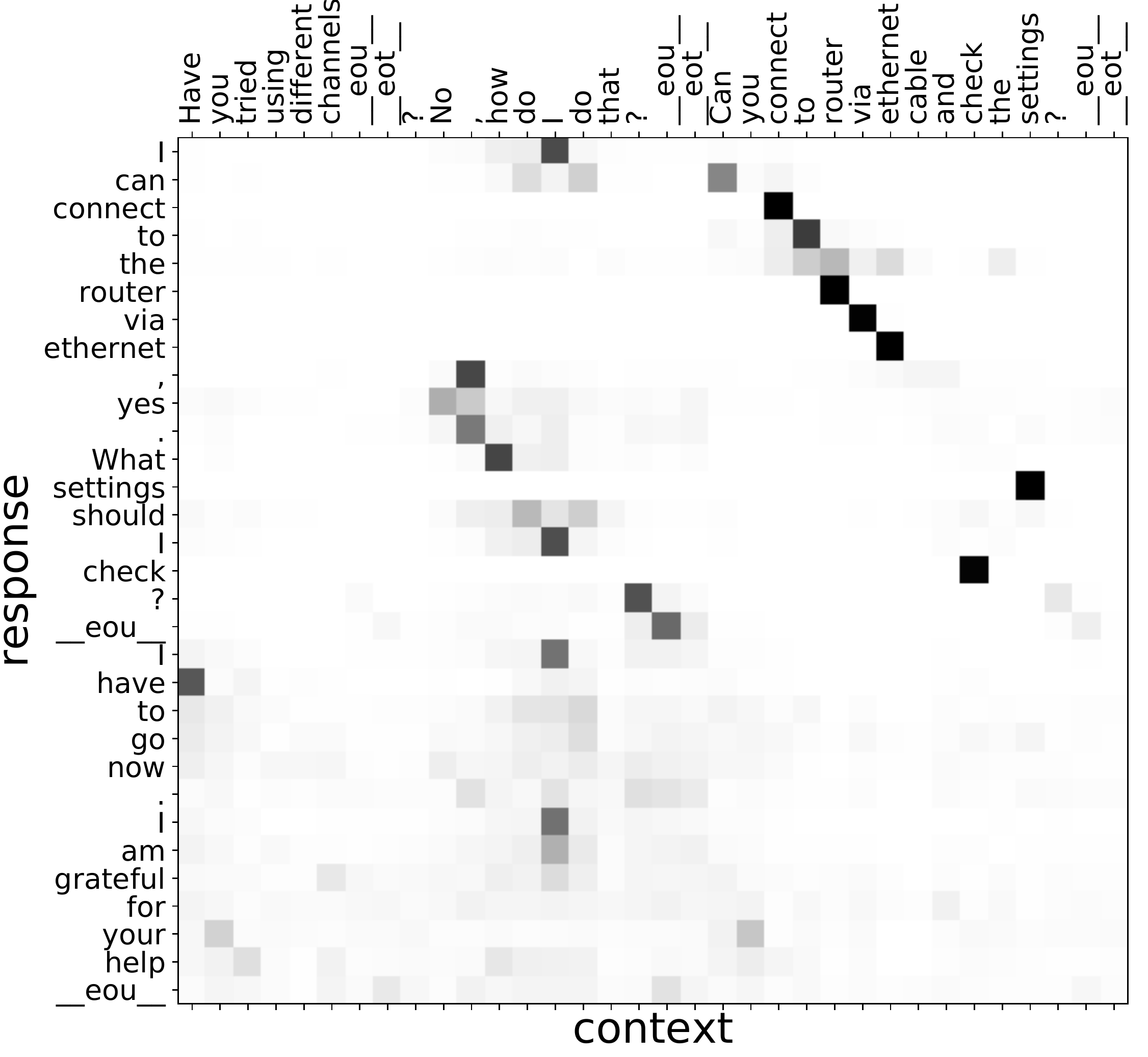}
    \caption{Response-to-context attention weights for a sample. The darker units mean larger values.}
    \label{fig3}
    \end{figure}

    \paragraph{Match} %The matching layer in IMN was ablated so that the outputs of the AHRE were followed by a pooling operation to directly obtain the sequence of utterance embeddings and the response embeddings.
    %The decreased performance indicates that bidirectional interactions between contexts and responses are beneficial for collecting matching information.
    The decreased performance indicates that interactions between contexts and responses are beneficial for matching.
    We conduct a case study and visualize the response-to-context weights used in Eq.~\ref{equ3} to demonstrate their ability to select relevant parts as shown in Figure~\ref{fig3}. Some important words such as ``\emph{connect}", ``\emph{router}" and ``\emph{ethernet}" in the response can select their relevant words in the context, and some unimportant words such as ``\emph{tried}", ``\emph{channels}" and ``\emph{the}" in the context occupy small weights when forming representations.

\section{Conclusion}
  In this paper, we propose an interactive matching network for the response selection task. An empirical study on four public datasets shows that our proposed model outperforms the baseline models on all metrics, achieving new state-of-the-art performance and showing compatibility across domains for multi-turn response selection.

\section*{Acknowledgements}
  We thank anonymous reviewers for their valuable comments.

%
% The next two lines define the bibliography style to be used, and the bibliography file.
\bibliographystyle{ACM-Reference-Format}
\bibliography{sample-base}


\begin{thebibliography}{13}

%%% the Web address.

\ifx \showCODEN    \undefined \def \showCODEN     #1{\unskip}     \fi
\ifx \showDOI      \undefined \def \showDOI       #1{#1}\fi
\ifx \showISBNx    \undefined \def \showISBNx     #1{\unskip}     \fi
\ifx \showISBNxiii \undefined \def \showISBNxiii  #1{\unskip}     \fi
\ifx \showISSN     \undefined \def \showISSN      #1{\unskip}     \fi
\ifx \showLCCN     \undefined \def \showLCCN      #1{\unskip}     \fi
\ifx \shownote     \undefined \def \shownote      #1{#1}          \fi
\ifx \showarticletitle \undefined \def \showarticletitle #1{#1}   \fi
\ifx \showURL      \undefined \def \showURL       {\relax}        \fi
% invisible to TeX
\providecommand\bibfield[2]{#2}
\providecommand\bibinfo[2]{#2}
\providecommand\natexlab[1]{#1}
\providecommand\showeprint[2][]{arXiv:#2}

\bibitem[\protect\citeauthoryear{Chen, Liu, Yin, and Tang}{Chen
  et~al\mbox{.}}{2017}]%
        {DBLP:journals/sigkdd/ChenLYT17}
\bibfield{author}{\bibinfo{person}{Hongshen Chen}, \bibinfo{person}{Xiaorui
  Liu}, \bibinfo{person}{Dawei Yin}, {and} \bibinfo{person}{Jiliang Tang}.}
  \bibinfo{year}{2017}\natexlab{}.
\newblock \showarticletitle{A Survey on Dialogue Systems: Recent Advances and
  New Frontiers}.
\newblock \bibinfo{journal}{\emph{{SIGKDD} Explorations}} \bibinfo{volume}{19},
  \bibinfo{number}{2} (\bibinfo{year}{2017}), \bibinfo{pages}{25--35}.
\newblock


\bibitem[\protect\citeauthoryear{Chen, Zhu, Ling, Wei, Jiang, and Inkpen}{Chen
  et~al\mbox{.}}{2016}]%
        {DBLP:conf/acl/ChenZLWJI17}
\bibfield{author}{\bibinfo{person}{Qian Chen}, \bibinfo{person}{Xiaodan Zhu},
  \bibinfo{person}{Zhenhua Ling}, \bibinfo{person}{Si Wei},
  \bibinfo{person}{Hui Jiang}, {and} \bibinfo{person}{Diana Inkpen}.}
  \bibinfo{year}{2016}\natexlab{}.
\newblock \showarticletitle{Enhanced lstm for natural language inference}.
\newblock \bibinfo{journal}{\emph{arXiv preprint arXiv:1609.06038}}
  (\bibinfo{year}{2016}).
\newblock


\bibitem[\protect\citeauthoryear{Hochreiter and Schmidhuber}{Hochreiter and
  Schmidhuber}{1997}]%
        {DBLP:journals/neco/HochreiterS97}
\bibfield{author}{\bibinfo{person}{Sepp Hochreiter} {and}
  \bibinfo{person}{J{\"{u}}rgen Schmidhuber}.} \bibinfo{year}{1997}\natexlab{}.
\newblock \showarticletitle{Long Short-Term Memory}.
\newblock \bibinfo{journal}{\emph{Neural Computation}} \bibinfo{volume}{9},
  \bibinfo{number}{8} (\bibinfo{year}{1997}), \bibinfo{pages}{1735--1780}.
\newblock


\bibitem[\protect\citeauthoryear{Lowe, Pow, Serban, and Pineau}{Lowe
  et~al\mbox{.}}{2015}]%
        {DBLP:conf/sigdial/LowePSP15}
\bibfield{author}{\bibinfo{person}{Ryan Lowe}, \bibinfo{person}{Nissan Pow},
  \bibinfo{person}{Iulian Serban}, {and} \bibinfo{person}{Joelle Pineau}.}
  \bibinfo{year}{2015}\natexlab{}.
\newblock \showarticletitle{The ubuntu dialogue corpus: A large dataset for
  research in unstructured multi-turn dialogue systems}.
\newblock \bibinfo{journal}{\emph{arXiv preprint arXiv:1506.08909}}
  (\bibinfo{year}{2015}).
\newblock


\bibitem[\protect\citeauthoryear{Lowe, Pow, Serban, Charlin, Liu, and
  Pineau}{Lowe et~al\mbox{.}}{2017}]%
        {DBLP:journals/dad/LowePSCLP17}
\bibfield{author}{\bibinfo{person}{Ryan~Thomas Lowe}, \bibinfo{person}{Nissan
  Pow}, \bibinfo{person}{Iulian~Vlad Serban}, \bibinfo{person}{Laurent
  Charlin}, \bibinfo{person}{Chia-Wei Liu}, {and} \bibinfo{person}{Joelle
  Pineau}.} \bibinfo{year}{2017}\natexlab{}.
\newblock \showarticletitle{Training end-to-end dialogue systems with the
  ubuntu dialogue corpus}.
\newblock \bibinfo{journal}{\emph{Dialogue \& Discourse}} \bibinfo{volume}{8},
  \bibinfo{number}{1} (\bibinfo{year}{2017}), \bibinfo{pages}{31--65}.
\newblock


\bibitem[\protect\citeauthoryear{Peters, Neumann, Iyyer, Gardner, Clark, Lee,
  and Zettlemoyer}{Peters et~al\mbox{.}}{2018}]%
        {DBLP:conf/naacl/PetersNIGCLZ18}
\bibfield{author}{\bibinfo{person}{Matthew~E Peters}, \bibinfo{person}{Mark
  Neumann}, \bibinfo{person}{Mohit Iyyer}, \bibinfo{person}{Matt Gardner},
  \bibinfo{person}{Christopher Clark}, \bibinfo{person}{Kenton Lee}, {and}
  \bibinfo{person}{Luke Zettlemoyer}.} \bibinfo{year}{2018}\natexlab{}.
\newblock \showarticletitle{Deep contextualized word representations}. In
  \bibinfo{booktitle}{\emph{Proceedings of NAACL-HLT}}.
  \bibinfo{pages}{2227--2237}.
\newblock


\bibitem[\protect\citeauthoryear{Wang and Jiang}{Wang and Jiang}{2016}]%
        {DBLP:journals/corr/WangJ16b}
\bibfield{author}{\bibinfo{person}{Shuohang Wang} {and} \bibinfo{person}{Jing
  Jiang}.} \bibinfo{year}{2016}\natexlab{}.
\newblock \showarticletitle{A Compare-Aggregate Model for Matching Text
  Sequences}.
\newblock \bibinfo{journal}{\emph{CoRR}}  \bibinfo{volume}{abs/1611.01747}
  (\bibinfo{year}{2016}).
\newblock


\bibitem[\protect\citeauthoryear{Wang, Hamza, and Florian}{Wang
  et~al\mbox{.}}{2017}]%
        {DBLP:conf/ijcai/WangHF17}
\bibfield{author}{\bibinfo{person}{Zhiguo Wang}, \bibinfo{person}{Wael Hamza},
  {and} \bibinfo{person}{Radu Florian}.} \bibinfo{year}{2017}\natexlab{}.
\newblock \showarticletitle{Bilateral multi-perspective matching for natural
  language sentences}. In \bibinfo{booktitle}{\emph{Proceedings of the 26th
  International Joint Conference on Artificial Intelligence}}. AAAI Press,
  \bibinfo{pages}{4144--4150}.
\newblock


\bibitem[\protect\citeauthoryear{Wu, Wu, Xing, Zhou, and Li}{Wu
  et~al\mbox{.}}{2017}]%
        {DBLP:conf/acl/WuWXZL17}
\bibfield{author}{\bibinfo{person}{Yu Wu}, \bibinfo{person}{Wei Wu},
  \bibinfo{person}{Chen Xing}, \bibinfo{person}{Ming Zhou}, {and}
  \bibinfo{person}{Zhoujun Li}.} \bibinfo{year}{2017}\natexlab{}.
\newblock \showarticletitle{Sequential Matching Network: A New Architecture for
  Multi-turn Response Selection in Retrieval-Based Chatbots}. In
  \bibinfo{booktitle}{\emph{Proceedings of the 55th Annual Meeting of the
  Association for Computational Linguistics (Volume 1: Long Papers)}}.
  \bibinfo{pages}{496--505}.
\newblock


\bibitem[\protect\citeauthoryear{Xu, Liu, Wang, Sun, and Wang}{Xu
  et~al\mbox{.}}{2016}]%
        {DBLP:journals/corr/XuLWSW16}
\bibfield{author}{\bibinfo{person}{Zhen Xu}, \bibinfo{person}{Bingquan Liu},
  \bibinfo{person}{Baoxun Wang}, \bibinfo{person}{Chengjie Sun}, {and}
  \bibinfo{person}{Xiaolong Wang}.} \bibinfo{year}{2016}\natexlab{}.
\newblock \showarticletitle{Incorporating loose-structured knowledge into lstm
  with recall gate for conversation modeling}.
\newblock \bibinfo{journal}{\emph{arXiv preprint arXiv:1605.05110}}
  \bibinfo{volume}{3} (\bibinfo{year}{2016}).
\newblock


\bibitem[\protect\citeauthoryear{Yoon, Shin, and Jung}{Yoon
  et~al\mbox{.}}{2018}]%
        {DBLP:conf/naacl/YoonSJ18}
\bibfield{author}{\bibinfo{person}{Seunghyun Yoon}, \bibinfo{person}{Joongbo
  Shin}, {and} \bibinfo{person}{Kyomin Jung}.} \bibinfo{year}{2018}\natexlab{}.
\newblock \showarticletitle{Learning to Rank Question-Answer Pairs using
  Hierarchical Recurrent Encoder with Latent Topic Clustering}. In
  \bibinfo{booktitle}{\emph{Proceedings of NAACL-HLT}}.
  \bibinfo{pages}{1575--1584}.
\newblock


\bibitem[\protect\citeauthoryear{Zhang, Li, Zhu, Zhao, and Liu}{Zhang
  et~al\mbox{.}}{2018}]%
        {DBLP:conf/coling/ZhangLZZL18}
\bibfield{author}{\bibinfo{person}{Zhuosheng Zhang}, \bibinfo{person}{Jiangtong
  Li}, \bibinfo{person}{Pengfei Zhu}, \bibinfo{person}{Hai Zhao}, {and}
  \bibinfo{person}{Gongshen Liu}.} \bibinfo{year}{2018}\natexlab{}.
\newblock \showarticletitle{Modeling Multi-turn Conversation with Deep
  Utterance Aggregation}. In \bibinfo{booktitle}{\emph{Proceedings of the 27th
  International Conference on Computational Linguistics}}.
  \bibinfo{pages}{3740--3752}.
\newblock


\bibitem[\protect\citeauthoryear{Zhou, Li, Dong, Liu, Chen, Zhao, Yu, and
  Wu}{Zhou et~al\mbox{.}}{[n. d.]}]%
        {DBLP:conf/acl/WuLCZDYZL18}
\bibfield{author}{\bibinfo{person}{Xiangyang Zhou}, \bibinfo{person}{Lu Li},
  \bibinfo{person}{Daxiang Dong}, \bibinfo{person}{Yi Liu},
  \bibinfo{person}{Ying Chen}, \bibinfo{person}{Wayne~Xin Zhao},
  \bibinfo{person}{Dianhai Yu}, {and} \bibinfo{person}{Hua Wu}.}
  \bibinfo{year}{[n. d.]}\natexlab{}.
\newblock \showarticletitle{Multi-turn response selection for chatbots with
  deep attention matching network}. In \bibinfo{booktitle}{\emph{Proceedings of
  the 56th Annual Meeting of the Association for Computational Linguistics}}.
\newblock


\end{thebibliography}

\appendix

\section{Supplemental Material}

  \subsection{Detailed Dataset Descriptions}

    We tested IMN on two English public multi-turn response selection datasets, Ubuntu Dialogue Corpus V1 \cite{DBLP:conf/sigdial/LowePSP15} and Ubuntu Dialogue Corpus V2 \cite{DBLP:journals/dad/LowePSCLP17}, and two Chinese datasets, Douban Conversation Corpus \cite{DBLP:conf/acl/WuWXZL17} and E-commerce Dialogue Corpus \cite{DBLP:conf/coling/ZhangLZZL18}.
    Ubuntu Dialogue Corpus V1 and V2 contain multi-turn dialogues about Ubuntu system troubleshooting in English.
    Here, we adopted the version of Ubuntu Dialogue Corpus V1 shared in \citet{DBLP:journals/corr/XuLWSW16}, in which numbers, paths and URLs were replaced by placeholders.
    Compared with Ubuntu Dialogue Corpus V1, the training, validation and test dialogues in the V2 dataset were generated in different periods without overlap. Besides, the V2 dataset discriminates between the end of an utterance (\_eou\_) and the end of a turn (\_eot\_). In both of the Ubuntu corpora, the positive responses are true responses from humans, and the negative responses are randomly sampled.
    The Douban Conversation Corpus was crawled from a Chinese social network on open-domain topics. It was constructed in a similar way to the Ubuntu corpus. The Douban Conversation Corpus collected responses via a small inverted-index system, and labels were manually annotated. The E-commerce Dialogue Corpus collected real-world conversations between customers and customer service staff from the largest e-commerce platform in China.
    Some statistics of these datasets are provided in Table~\ref{tab1}.

    \begin{table}[!hbt]
   %\small
   \caption{Statistics of the datasets that our model is tested on.}
   \centering
   \setlength{\tabcolsep}{1.6mm}{
   \begin{tabular}{c|c|ccc}
   \toprule
    \multicolumn{2}{c|}{Dataset}                      & Train & Valid & Test \\
    \hline
    \multirow{3}*{Ubuntu V1}    & pairs               & 1M    & 0.5M  & 0.5M \\
                                & positive:negative   & 1: 1  & 1: 9  & 1: 9 \\
                                & positive/context    & 1     & 1     & 1    \\
    \hline
    \multirow{3}*{Ubuntu V2}    & pairs               & 1M    & 195k  & 189k \\
                                & positive:negative   & 1: 1  & 1: 9  & 1: 9 \\
                                & positive/context    & 1     & 1     & 1    \\
    \hline
    \multirow{3}*{Douban}       & pairs               & 1M    & 50k   & 10k  \\
                                & positive:negative   & 1: 1  & 1: 1  & 1: 9 \\
                                & positive/context    & 1     & 1     & 1.18 \\
    \hline
    \multirow{3}*{E-commerce}   & pairs               & 1M    & 10k   & 10k  \\
                                & positive:negative   & 1: 1  & 1: 1  & 1: 9 \\
                                & positive/context    & 1     & 1     & 1    \\
   \bottomrule
   \end{tabular}}
   \label{tab1}
   \end{table}

  \subsection{Training Details}

    The Adam method was employed for optimization, with a batch size of 96 for the two English datasets and 128 for the two Chinese datasets.
    The initial learning rate was 0.001 and was exponentially decayed by 0.96 every 5000 steps.
    Dropout with a rate of 0.2 was applied to the word embeddings and all hidden layers.

    The word embeddings for the English datasets were concatenations of the 300-dimensional GloVe embeddings, 100-dimensional embeddings estimated on the training set using the Word2Vec algorithm and 150-dimensional character-level embeddings with window sizes of \{3, 4, and 5\}, each consisting of 50 filters.
    The word embeddings for the Chinese datasets were concatenations of the 200-dimensional embeddings from and the 200-dimensional embeddings estimated on the training set using the Word2Vec algorithm. Character-level embeddings were not employed for the two Chinese datasets due to the large number of Chinese characters. The word embeddings were not updated during training.

    All hidden states of the LSTM had 200 dimensions. The number of BiLSTM layers in the AHRE was 3.
    The MLP at the prediction layer had a hidden unit size of 256 with ReLU activation.
    The maximum word length was set to 18, the maximum utterance length was set to 50, and the maximum context length was set to 10. We padded with zeros if the number of utterances in a context was less than 10; otherwise, we kept the last 10 utterances.
    We used the development dataset to set the stop condition to select the best model for testing.

    %All codes were implemented in the TensorFlow framework and will be published to help replicate our results after paper acceptance.%\footnote{https://github.com/JasonForJoy/IMN}.

  \subsection{Performance of Different Number of Layers in AHRE}
\begin{figure}
    \centering
    \includegraphics[width=6cm]{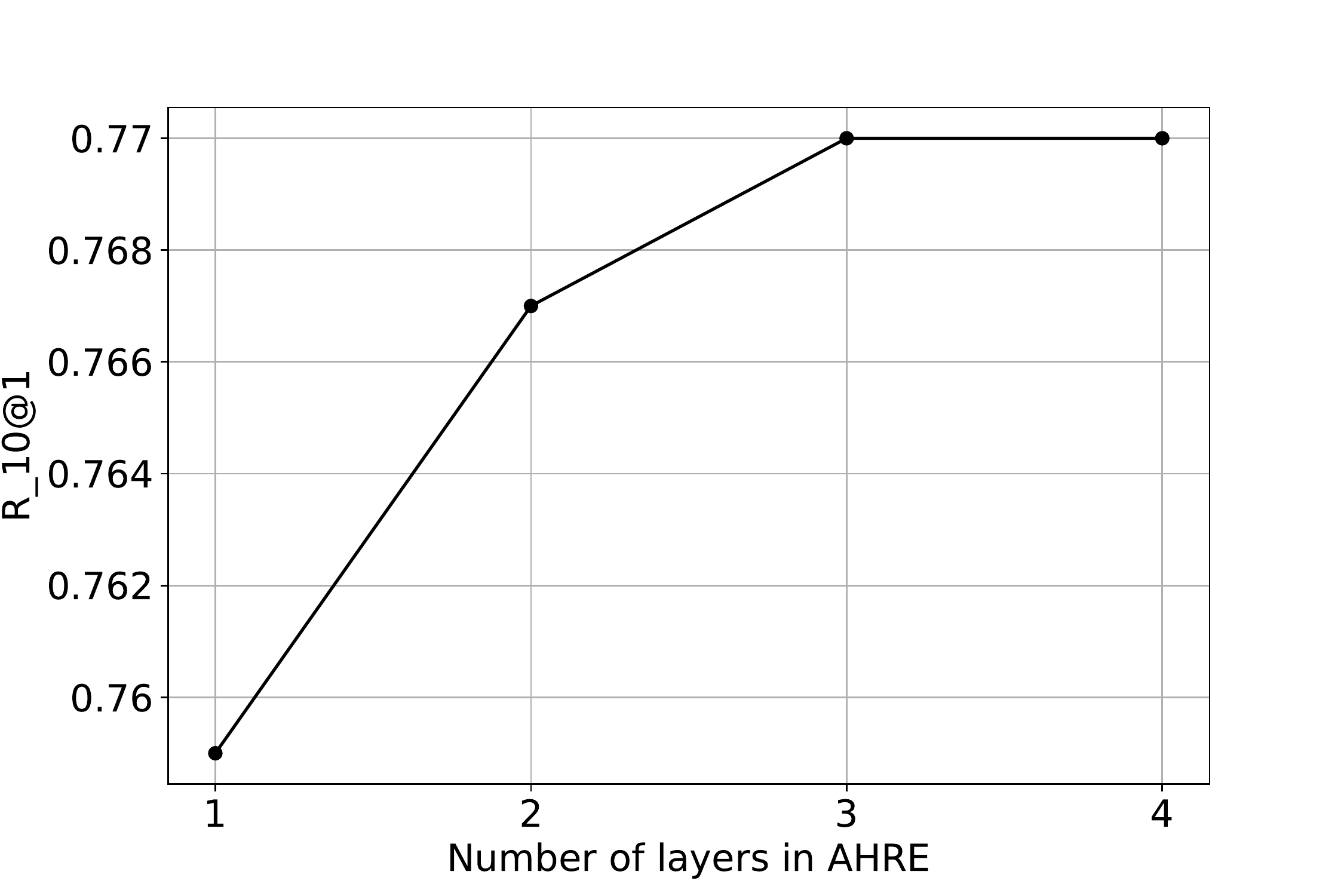}
    \caption{Performance of IMN for different numbers of layers in the AHRE on Ubuntu V2 validation set.}
    \label{fig2}
    \end{figure}

    For the Ubuntu V2 dataset, the number of layers in the AHRE was tuned on its validation set. Using three layers achieved the best performance as shown in Figure~\ref{fig2}.

\end{document}